\documentclass[a4]{article}
\usepackage{graphicx}
\usepackage{algorithm2e}
\usepackage{amssymb}
\usepackage{hyperref}

\def\be{\begin{equation}}
\def\ee{\end{equation}}
\newcommand{\ff}[1]{{\bf  #1}}
\def\a{\alpha}

\def\lam{\lambda}

\def\x{\ff{x}}
\def\v{\ff{v}}
\def\ra{\rightarrow}

\def\x{\ff{x}}

\begin{document}

\title{Social Algorithms}

\author{Xin-She Yang \\[10pt]
School of Science and Technology, Middlesex University, London, UK. }

\date{}

\maketitle

\noindent {\bf Citation Details}: 
X.-S. Yang, Social Algorithms, in: {\it Encyclopedia of Complexity and Systems Science} (Edited by R. A. Meyers), Springer, (2017). 
\href{https://doi.org/10.1007/978-3-642-27737-5_678-1}{https://doi.org/10.1007/978-3-642-27737-5\_678-1}



\section*{Glossary} \href{https://link.springer.com/referenceworkentry/10.1007%2F978-3-642-27737-5_678-1}{Social Algorithms}

\begin{itemize}

\item {\bf Algorithm}: An algorithm is a step-by-step, computational procedure or
a set of rules to be followed by a computer in calculations or computing
an answer to a problem. \index{algorithm} \index{ant colony optimization}

\item {\bf Ant colony optimization}: Ant colony optimization (ACO) is an algorithm
for solving optimization problems such as routing problems using multiple
agents. ACO mimics the local interactions of social ant colonies and the
use of chemical messenger -- pheromone to mark paths. No centralized control
is used and the system evolves according to simple local interaction rules.
\index{bat algorithm}

\item {\bf Bat algorithm}: Bat algorithm (BA) is an algorithm for optimization, which uses frequency-tuning to mimic the basic behaviour of echolocation of microbats. BA also uses the variations of loudness and pulse emission rates and a solution vector to a problem corresponds to a position vector of a bat in the search space. Evolution of solutions follow two algorithmic equations for positions and frequencies.

\item {\bf Bees-inspired algorithms}: Bees-inspired algorithms are a class of algorithms for optimization using the foraging characteristics of honeybees and their labour division to carry out search. Pheromone may also be used in some variants of bees-inspired algorithms.

    \index{bees-inspired algorithm} \index{cuckoo search}
\item {\bf Cuckoo Search}: Cuckoo search (CS) is an optimization algorithm that mimics the brood parasitism of some cuckoo species. A solution to a problem is considered as an egg laid by a cuckoo. The evolution of solutions is carried out by L\'evy flights and the similarity of solutions controlled by a switch probability. \index{firefly algorithm}

\item {\bf Firefly algorithm}: Firefly algorithm is an optimization inspired by the flashing patterns of tropical fireflies. The location of a firefly is equivalent to a solution vector to a problem, and the evolution of fireflies follows a nonlinear equation to simulate the attraction between fireflies of different brightness that is linked to the objective landscape of the problem.

      \index{metaheuristic}
\item {\bf Metaheuristic}: Metaheuristic or metaheuristic algorithms are a class of optimization algorithms designed by drawing inspiration from nature. They are thus mostly nature-inspired algorithms, and examples of such metaheuristic algorithms are ant colony optimization, firefly algorithm, and particle swarm optimization. These algorithms are often swarm intelligence based algorithms.

       \index{nature-inspired computation}
\item {\bf Nature-Inspired computation}: Nature-inspired computation is an area of computer science, concerning the development and application of nature-inspired metaheuristic algorithms for optimization, data mining, machine learning and computational intelligence.

\item {\bf Nature-inspired algorithms}: Nature-inspired algorithms are a much wider class of algorithms that have been developed by drawing inspiration from nature. These algorithms are almost all population-based algorithms. For example, ant colony optimization, bat algorithm, cuckoo search and particle swarm optimization are all nature-inspired algorithms.

\item {\bf Objective}: An objective function is the function to be optimized in an optimization problem. Objective functions are also called cost functions, loss functions, utility functions, or fitness functions. \index{objective} \index{optimization}

\item {\bf Optimization}: Optimization concerns a broad area in mathematics,
computer science,
operations research and engineering designs. For example, mathematical programming or mathematical optimization is traditionally an integrated part of operations research. Nowadays, optimization is relevant to almost every area of sciences and engineering. Optimization problems are formulated with one or more objective functions subject to various constraints. Objectives can be either minimized or maximized, depending on the formulations. Optimization can subdivide into linear programming and nonlinear programming. \index{particle swarm optimization}

\item {\bf Particle swarm optimization}: Particle swarm optimization (PSO) is an optimization algorithm that mimics the basic swarming behaviour of fish and birds. Each particle has a velocity and a position that corresponds to a solution to a problem. The evolution of the particles is governed by two equations with the use of the best solution found in the swarm.

\item {\bf Population-based algorithm}: A population-based algorithm is an algorithm using a group of multiple agents such as particles, ants and fireflies to carry out search for optimal solutions. The initialization of the population is usually done randomly and the evolution of the population is governed by the main governing equations in an algorithm in an iterative manner. All social algorithms are population-based algorithms.

   \index{social algorithm}
\item {\bf Social Algorithms}: Social algorithms are a class of nature-inspired algorithms that use some of characteristics of social swarms such as social insects (e.g., ants, bees, bats and fireflies) and reproduction strategies such as cuckoo-host species co-evolution. These algorithms tend to be swarm intelligence based algorithms. Examples are ant colony optimization, particle swarm optimization and cuckoo search.

    \index{swarm intelligence}
\item {\bf Swarm intelligence}: Swarm intelligence is the emerging behaviour of multi-agent systems where multiple agents interact and exchange information, according to simple local rules. There is no centralized control, and each agent follows local rules such as following pheromone trail and deposit pheromone. These rules can often expressed as simple dynamic equations and the system then evolves iteratively. Under certain conditions, emergent behaviour such as self-organization may occur, and the system may show higher-level structures or behaviour that is often more complex than that of individuals.

\end{itemize}

\section{Introduction}

To find solutions to problems commonly used in science and engineering, algorithms are required.
An algorithm is a step-by-step computational procedure or a set of rules to be followed by a computer. One of the oldest algorithms is the Euclidean algorithm for finding the greatest common divisor (gcd) of two integers such as 12345 and 125, and this algorithm was first given in detail in Euclid's Elements about 2300 years ago (Chabert 1999).  Modern computing involves a large set of different algorithms from fast Fourier transform (FFT) to image processing techniques and from conjugate gradient methods to finite element methods.

Optimization problems in particular require specialized optimization techniques, ranging from the simple Newton-Raphson's method to more sophisticated simplex methods for linear programming.
Modern trends tend to use a combination of traditional techniques in combination with contemporary stochastic metaheuristic algorithms such as genetic algorithms, firefly algorithm and particle swarm optimization.

This work concerns a special class of algorithms for solving optimization problems and these algorithms fall into a category: social algorithms, which can in turn belong to swarm intelligence in general. Social algorithms use multiple agents and the `social' interactions to design rules for algorithms so that such social algorithms can mimic certain successful characteristics of the social/biological systems such as ants, bees, birds and animals. Therefore, our focus will solely be on such social algorithms.

It is worth pointing out that the social algorithms in the present context do not include the algorithms for social media, even though algorithms for social media analysis are sometimes simply referred to as `social algorithm' (Lazer 2015). The social algorithms in this work are mainly nature-inspired, population-based algorithms for optimization, which share many similarities with swarm intelligence.

Social algorithms belong to a wider class of metaheuristic algorithms. Alan Turing, the pioneer of artificial intelligence, was the first to use heuristic in his Enigma-decoding work during the Second World War and connectionism (the essence of neural networks) as outlined in his National Physical Laboratory report {\it Intelligent Machinery} (Turing 1948).
\index{intelligent machinery}

The initiation of non-deterministic algorithms was in the 1960s when evolutionary strategy and genetic algorithm started to appear, which attempted to simulate the key feature of Darwinian evolution of biological systems. \index{evolutionary strategy} \index{genetic algorithm}
For example, genetic algorithm (GA) was developed by John Holland in the 1960s (Holland, 1975), which uses crossover, mutation and selection as basic genetic operators for algorithm operations. At about the same period, Ingo Recehberg and H. P. Schwefel, developed the evolutionary strategy for constructing automatic experimenter using simple rules of mutation and selection, though crossover was not used. In around 1966, L. J. Fogel and colleagues used simulated evolution as a learning tool to study artificial intelligence, which leads to the development of evolutionary programming. All these algorithms now evolved into a much wider discipline, called evolutionary algorithms or evolutionary computation (Fogel et al. 1966).
\index{simulated annealing}

Then, simulated annealing was developed in 1983 by Kirpatrick et al. (1983) which simulated the annealing process of metals for the optimization purpose, and the Tabu search was developed by Fred Glover in 1986 (Glover 1986) that uses memory and history to enhance the search efficiency. In fact, it was Fred Glover who coined the word `metaheuristic' in his 1986 paper. \index{metaheuristic}

The major development in the context of social algorithms started in the 1990s. First, Marco Dorigo developed the ant colony optimization (ACO) in his PhD work (Dorigo 1992), and ACO uses the key characteristics of social ants to design procedure for optimization. Local interactions using pheromone and rules are used in ACO. \index{ant colony optimization} \index{particle swarm optimization}
Then, in 1995, particle swarm optimization was developed by James Kennedy and Russell C. Eberhardt, inspired by the swarming behaviour of fish and birds (Kennedy and Eberhart 1995). Though developed in 1997, differential evolution (DE) is not a social algorithm; however, DE has use vectorized mutation which forms a basis for many later algorithms (Storn and Price 1997). \index{differential evolution}

\index{no-free-lunch theorem}
Another interesting development is that no-free-lunch (NFL) theorems was proved in 1997 by D.H. Wolpert and W.G. Macready, which had much impact in the optimization and machine learning communities (Wolpert and Macready 1997). This basically dashed the dreams for finding the best algorithms for all problems because NFL theorems state that all algorithms are equally effective if measured in terms of averaged performance for {\it all} possible problems. Then, researchers realized that the performance and efficiency in practice are not measured by averaging over all possible problems. Instead, we are more concerned with a particular class of problems in a particular discipline, and there is no need to use an algorithm to solve all possible problems. Consequently, for a finite set of problems and for a given few algorithms, empirical observations and experience suggest that some algorithms can perform better than others. For example, algorithms that can use problem-specific knowledge such as convexity can be more efficient than random search. Therefore, further research should identify the types of problems that a given algorithm can solve, or the most suitable algorithms for a given type of problems. Thus, research resumes and continue, just with a different emphasis and from different perspectives.

At the turn of this century, developments of social algorithms became more active. In 2004, a honeybee algorithm for optimizing Internet hosting centres was developed by Sunil Nakrani and Craig Tovey (Nakrani and Tovey, 2004). In 2005, Pham et al. (2005) developed the bees algorithm, and the virtual bee algorithm was developed by Xin-She Yang in 2005 (Yang  2005). About the same time, the artificial bee colony (ABC) algorithm was developed by D. Karaboga in 2005 (Karaboga 2005). All these algorithms are bee-based algorithms and they all use some (but different) aspects of the foraging behaviour of social bees. \index{bees-inspired algorithm}

\index{firefly algorithm}
Then, in late 2007 and early 2008, the firefly algorithm (FA) was developed by Xin-She Yang, inspired by the flashing behaviour of tropic firefly species (Yang 2008). The attraction mechanism, together with the variation of light intensity, was used to produce a nonlinear algorithm that can deal with multiomodal optimization problems. In 2009, cuckoo search (CS) was developed by Xin-She Yang and Suash Deb, inspired by the brood parasitism of the reproduction strategies of some cuckoo species (Yang and Deb 2009). \index{cuckoo search}
This algorithm simulated partly the complex social interactions of cuckoo-host species co-evolution. Then, in 2010, the bat algorithm (BA) was developed by Xin-She Yang, inspired by the echolocation characterisitics of microbats (Yang 2010b), \index{bat algorithm}
and uses frequency-tuning in combination with the variations of loudness and pulse emission rates during foraging.  All these algorithms are can be considered as social algorithms because they use the `social' interactions and their biologically-inspired rules.

There are other algorithms developed in the last two decades, but they are not social algorithms.
For example,  harmony search is a music-inspired algorithm  (Geem et al. 2001), while gravitational search algorithm (GSA) is a physics-inspired algorithm (Rashedi et al. 2009). In addition, flower pollination algorithm (FPA) is an algorithm inspired by the pollination features of flowering plants (Yang 2012) with promising applications (Yang et al. 2014, Rodrigues et al. 2016). \index{flower pollination algorithm}
All these algorithms are population-based algorithm, but they do not strictly belong in swarm intelligence based or social algorithms. A wider range of applications of nature-inspired algorithms can be found in the recent literature (Yang et al. 2015, Yang and Papa 2016, Yang 2018).

As the focus of this work is on social algorithms, we will now explain some of the social algorithms in greater details.

\section{Algorithms and Optimization}

\index{algorithm} \index{optimization}
In order to demonstrate the role of social algorithms in solving optimization problems, let us first briefly outline the essence of an algorithm and the general formulation of an optimization problem.

\subsection{Essence of an Algorithm}
An algorithm is a computational, iterative procedure. For example, Newton's method for finding the roots of
a polynomial $p(x)=0$ can be written as
\be x_{t+1}=x_t -\frac{p(x_t)}{p'(x_t)}, \ee
where $x_t$ is the approximation at iteration $t$, and $p'(x)$ is the first derivative of $p(x)$.
This procedure typically starts with an initial guess $x_0$ at $t=0$.

In most cases, as along as $p' \ne 0$ and $x_0$ is not too far away from the target solution, this algorithm can work very well. As we do not know the target solution $x_*=\lim_{t \ra \infty} x_t$ in advance, the initial guess can be an educated guess or a purely random guess. However, if the
initial guess is too far way, the algorithm may never reach the final solution or simply fail.
For example, for $p(x)=x^2+9x-10=(x-1)(x+10)$, we know its roots are $x_*=1$ and $x_*=-10$.
We also have $p'(x)=2x+9$ and
 \be x_{t+1}=x_t -\frac{(x_t^2+9 x_t -10)}{2 x_t+9}. \ee
 If we start from $x_0=10$, we can easily reach $x_*=1$ in less than 5 iterations. If we use $x_0=100$, it may take about 8 iterations, depending on the accuracy we want. If we start any value $x_0>0$, we can only reach $x_*=1$ and we will never reach the other root $x_*=-10$. If we start with $x_0=-5$, we can reach $x_*=-10$  in about 7 steps with an accuracy of $10^{-9}$.  However, if we start with $x_0=-4.5$, the algorithm will simply fail because $p'(x_0)=2x_0+9=0$.

This has clearly demonstrated that the final solution will usually depend on where the initial solution is.

This method can be modified to solve optimization problems. For example, for a single objective function $f(x)$, the minimal and maximal values should occur at stationary points $f'(x)=0$, which becomes a root-finding problem for $f'(x)$. Thus, the maximum or minimum of $f(x)$ can be found by modifying the
Newton's method as the following iterative formula:\index{Newton's method}
\be x_{t+1}=x_t -\frac{f'(x_t)}{f''(x_t)}. \ee

For a $D$-dimensional problem with an objective $f(\x)$ with independent variables
$\x=(x_1, x_2, ..., x_D)$, the above iteration formula can be generalized to a vector form
\be \x^{t+1}=\x^t -\frac{\nabla f(\x^t)}{\nabla^2 f(\x^t)}, \ee
where we have used the notation convention $\x^t$ to denote the current solution vector
at iteration $t$ (not to be confused with an exponent).

In general, an algorithm $A$ can be written as
\be \x^{t+1}=A(\x^t, \x_*, p_1,..., p_K), \ee
which represents that fact that the new solution vector is a function of the
existing solution vector $\x^t$, some historical best solution $\x_*$ during the
iteration history and a set of
algorithm-dependent parameters $p_1, p_2, ..., p_K$. The exact function forms will
depend on the algorithm, and different algorithms are only different in terms of
the function form, number of parameters and the ways of using historical data.

\subsection{Optimization}
In general, an optimization problem can be formulated in a $D$-dimensional design space as
\be \textrm{minimize }\;\; f(\x), \quad \x=(x_1, x_2, ..., x_D) \in \mathbb{R}^D, \ee
subject to
\be h_i(\x) =0, \;\; (i=1,2,..., M), \quad g_j(\x) \le 0, \;\; (j=1,2,..., N), \ee
where $h_i$ and $g_j$ are the equality constraints and inequality constraints, respectively.
In a special case when the problem functions $f(\x)$, $h_i(\x)$ and $g_j(\x)$ are all
linear, the problem becomes linear programming, which can be solved efficiently by using
George Dantzig's Simplex Method. However, in most cases, the problem functions $f(\x)$, $h_i(\x)$ and $g_j(\x)$ are all nonlinear, and such nonlinear optimization problems can be challenging to solve. There are a wide class of optimization techniques, including linear programming, quadratic programming, convex optimization, interior-point method, trust-region method, conjugate-gradient methods (S\"uli and Mayer 2003, Yang 2010c) as well as
evolutionary algorithms (Goldberg 1989), heuristics (Judea 1984) and metaheuristics (Yang 2008, Yang 2014b).

An interesting way of looking at algorithms and optimization is to consider an algorithm system as a complex, self-organized system (Ashby 1962, Keller 2009), \index{self-organization}
but nowadays researchers tend to look at algorithms from the point of view of swarm intelligence (Kennedy et al. 2001, Engelbrecht 2005, Fisher 2009, Yang 2014b).
\index{swarm intelligence}

\subsection{Traditional Algorithms or Social Algorithms?}

\index{social algorithm}
As there are many traditional optimization techniques, a natural question is why we need new algorithms such as social algorithms? One may wonder what is wrong with traditional algorithms? A short answer is that there is nothing wrong. Extensive literature and studies have demonstrated that traditional algorithms work quite well for many different types of problems, but
they do have some serious drawbacks:

\begin{itemize}
\item Traditional algorithms are mostly local search, there is no guarantee for global optimality for most optimization problems, except for linear programming and convex optimization. Consequently, the final solution will often depend on the initial starting points (except for linear programming and convex optimization).

\item Traditional algorithms tend to be problem-specific because they usually use some information such as derivatives about the local objective landscape. They cannot solve highly nonlinear, multimodal problems effectively, and they struggle to cope with problems with discontinuity, especially when gradients are needed.

\item These algorithms are largely deterministic, and thus the exploitation ability is high, but their exploration ability and diversity of solutions are low.

\end{itemize}

Social algorithms, in contrast, attempts to avoid these disadvantages by using a population-based approach with non-deterministic or stochastic
components to enhance their exploration ability. Compared with traditional algorithms, metaheuristic social algorithm are mainly designed for global search and tend to have the following advantages and characteristics:
\begin{itemize}

\item Almost all social algorithms are global optimizers, it is more likely to find the true global optimality. They are usually gradient-free methods and they do not use any derivative information, and thus can deal with highly nonlinear problems and problems with discontinuities.

\item They often treat problems as a black box without specific knowledge, thus they can solve a wider range of problems.

\item Stochastic components in such algorithms can increase the exploration ability and also enable the algorithms to escape any local modes (thus avoiding being trapped locally). The final solutions tend to `forget' the starting points, and thus independent of any initial guess and incomplete knowledge of the problem under consideration.

\end{itemize}

Though with obvious advantages, social algorithms do have some disadvantages. For example,
the computational efforts of these algorithms tend to be higher than those for traditional algorithms because more iterations are needed. Due to the stochastic nature, the final solutions obtained by such algorithms cannot be repeated exactly, and multiple runs should be carried out
to ensure consistency and some meaningful statistical analysis.

\section{Social Algorithms}

The literature of social algorithms and swarm intelligence is expanding rapidly, here we will introduce some of the most recent and widely used social algorithms.

\subsection{Ant Colony Optimization}
\index{ant colony optimization}

Ants are social insects that live together in well-organized colonies with a population size ranging from about 2 million to 25 million. Ants communicate with each other and interact with their environment in a swarm using local rules and scent chemicals or pheromone. There is no centralized control. Such a complex system with local interactions can self-organize with emerging behaviour, leading to some form of social intelligence.

Based on these characteristics, the ant colony optimization (ACO) was developed by Marco Dorigo in 1992 (Dorigo 1992),
and ACO attempts to mimic the foraging behaviour of social ants in a colony.
Pheromone is deposited by each agent, and such chemical will also evaporate.
The model for pheromone deposition and evaporation may vary slightly, depend on
the variants of ACO. However, in most cases, incremental deposition and exponential
decay are used in the literature.

From the implementation point of view, for example, a solution in a network optimization problem can be a path or route. Ants will explore the network paths and deposit pheromone when it moves. The quality of a solution is related to the pheromone concentration on the path. At the same time, pheromone will evaporate as (pseudo)time increases. At a junction with multiple routes,
the probability of choosing a particular route is determined by a decision criterion, depending on the normalized concentration of the route, and relative fitness of this route, comparing with all others. For example, in most studies, the probability $p_{ij}$ of choose a route from node $i$ to node $j$ can be calculated by
\be p_{ij}=\frac{\phi_{ij}^{\alpha} d_{ij}^{\beta}}{\sum_{i,j}^n \phi_{ij}^{\alpha} d_{ij}^{\beta}}, \ee
where $\alpha,\beta>0$ are the so-called influence parameters, and $\phi_{ij}$ is the pheromone concentration on the route between $i$ and $j$.
In addition, $d_{ij}$ is the desirability of the route (for example, the distance of the overall path). In the simplest case when $\alpha=\beta=1$, the choice probability is simply proportional to the pheromone concentration.

It is worth pointing out that ACO is a mixed of procedure and some simple equations such as
pheromone deposition and evaporation as well as the path selection probability. ACO has been applied to many applications from scheduling to routing problems (Dorigo 1992).

\subsection{Particle Swarm Optimization}
\index{particle swarm optimization}

Many swarms in nature such as fish and birds can have higher-level behaviour, but they all
obey simple rules. For example, a swarm of birds such as starlings simply follow three basic rules:
each bird flies according to the flight velocities of their neighbour birds (usually about seven adjacent birds), and birds on the edge of the swarm tend to fly into the centre of the swarm (so as to avoid being eaten by potential preditors such as eagles). In addition, birds tend to fly to search for food or shelters, thus a short memory is used. Based on such swarming characteristics, particle swarm optimization (PSO) was developed by Kennedy and Eberhart in 1995, which uses equations to simulate the swarming characteristics of birds and fish (Kennedy and Eberhart 1995).

For the ease of discussions below, let us use $\x_i$ and $\v_i$ to denote the position (solution) and velocity, respectively,  of a particle or agent $i$. In PSO, there are $n$ particles as a population, thus $i=1,2,...,n$. There are two equations for updating positions and velocities of particles, and they can be written as follows:
\be \ff{v}_i^{t+1}= \ff{v}_i^t  + \alpha \ff{\epsilon}_1
[\ff{g}^*-\x_i^t] + \beta \ff{\epsilon}_2 [\x_i^*-\x_i^t],
\label{pso-speed-100}
\ee
\be \x_i^{t+1}=\x_i^t+\ff{v}_i^{t+1}, \label{pso-speed-140} \ee
where $\ff{\epsilon}_1$ and $\ff{\epsilon}_2$ are two uniformly distributed random numbers in [0,1]. The learning parameters $\alpha$ and $\beta$ are usually in the range of [0,2]. In the above equations, $\ff{g}^*$ is the best solution found so far by all the particles in the population,
and each particle has an individual best solution $\x_i^*$ by
itself during the entire past iteration history.

It is clearly seen that the above algorithmic equations are linear in the sense that both equation
only depends on $\x_i$ and $\v_i$ linearly. PSO has been applied in many applications, and it has been extended to solve multiobjective optimization problems (Kennedy et al. 2001, Engelbrecht 2005). However, there are some drawbacks because PSO can often have so-called premature convergence
when the population loses diversity and thus gets stuck locally. Consequently, there are more than 20 different variants to try to remedy this with various degrees of improvements.

\subsection{Bees-inspired Algorithms}
\index{bees-inspired algorithm}

Bees such as honeybees live a colony and there are many subspecies of bees. Honeybees have three castes, including worker bees, queens and drones. The division of labour among bees is interesting, and worker bees forage, clean hive and defense the colony, and they have to collect and store honey. Honeybees communicate by pheromone and `waggle dance' and other local interactions, depending on species. Based on the foraging and social interactions of honeybees, researchers have developed various forms and variants of bees-inspired algorithms.

The first use of bees-inspired algorithms was probably by S. Nakrani and C. A. Tovey in 2004 to study web-hosting servers (Nakrani and Tovey 2004), while slightly later in 2004 and early 2005, Yang used the virtual bee algorithm to solve optimization problems (Yang 2005). At around the same time, D. Karaboga used the artificial bee colony (ABC) algorithm to carry out numerical optimization (Karaboga 2005). In addition, Pham et al. (2005) used bees algorithm to solve continuous optimization and function optimization problems.
In about 2007, Afshar et al. (2007) used a honey-bee mating optimization approach for optimizing reservoir operations.

For example, in ABC, the bees are divided into three groups: forager bees, onlooker bees and scouts. For each food source, there is one forager bee who shares information with onlooker bees after returning to the colony from foraging, and the number of forager bees is equal to the number of food sources. Scout bees do random flight to explore, while a forager at a scarce food source may have to be forced to become a scout bee.  The generation of a new solution $v_{i,k}$ is done by
\be v_{i,k}=x_{i,k} + \phi (x_{i,k}-x_{j,k}), \ee
which is updated for each dimension $k=1,2, ..., D$ for different solutions (e.g., $i$ and $j$) in a population of $n$ bees ($i,j=1,2,...,n)$. Here, $\phi$ is a random number in [-1,1].  A food source is chosen by a roulette-based probability criterion, while a scout bee uses a Monte Carlo style randomization between the lower bound (L) and the upper bound (U).
\be x_{i,k}=L_k+ r (U_k-L_k), \ee
where $k=1,2,..., D$ and $r$ is a uniformly distributed random number in [0,1].

Bees-inspired algorithms have been applied in many applications with diverse characteristics and variants (Pham et al. 2005, Karaboga 2005).

\subsection{Bat Algorithm}
\index{bat algorithm}

Bats are the only mammals with wings, and it is estimated that there are about 1000 different bat species. Their sizes can range from tiny bumblebee bats to giant bats. Most bat species use echolocation to a certain degree, though microbats extensively use echolocation for foraging and navigation.
Microbats emit a series of loud, ultrasonic sound pules and listen their echoes to `see' their surrounding. The pulse properties vary and correlate with their hunting strategies. Depending on the species, pulse emission rates will increase when homing for prey with frequency-modulated short pulses
(thus varying wavelengths to increase the detection resolution). Each pusle may last about 5 to 20 milliseconds with a frequency range of 25 kHz to 150 kHz, and the spatial resolution can be as small as a few millimetres, comparable to the size of insects they hunt.

Bat algorithm (BA), developed by Xin-She Yang in 2010,
uses some characteristics of frequency-tuning and
echolocation of microbats (Yang 2010b, Yang 2011).
It also uses the variations of pulse emission rate $r$ and loudness $A$ to control exploration
and exploitation.  In the bat algorithm, main algorithmic equations for position $\x_i$ and velocity $\ff{v}_i$ for bat $i$ are
\be f_i =f_{\min} + (f_{\max}-f_{\min}) \beta, \label{f-equ-150} \ee
\be \ff{v}_i^{t} = \ff{v}_i^{t-1} +  (\x_i^{t-1} - \x_*) f_i , \ee
\be \x_i^{t}=\x_i^{t-1} + \ff{v}_i^t,  \label{f-equ-250} \ee
where $\beta \in [0,1]$ is a random vector drawn from a uniform distribution so that the frequency can vary from $f_{\min}$ to $f_{\max}$.
Here, $\x_*$ is the current best solution found so far by
all the virtual bats.

From the above equations, we can see that both equations are linear in terms of
$\x_i$ and $\v_i$.  But, the control of exploration and exploitation is carried out
by the variations of loudness $A(t)$ from a high value to a lower value and
the emission rate $r$ from a lower value to a higher value. That is
\be A_i^{t+1}=\a A_i^t, \quad r_i^{t+1}= r_i^0 (1-e^{-\gamma t}), \ee
where $0<\a<1$ and $\gamma>0$ are two parameters.
As a result, the actual
algorithm can have a weak nonlinearity. Consequently, BA can have a faster convergence rate
in comparison with PSO. BA has been extended to multiobjective optimization and hybrid versions (Yang 2011, Yang 2014b).

\subsection{Firefly Algorithm}
\index{firefly algorithm}

There are about 2000 species of fireflies and most species produce short, rhythmic flashes by bioluminescence. Each species can have different flashing
patterns and rhythms, and one of the main functions of such flashing light acts as a signaling system to communicate with other fireflies. As light intensity in the night sky decreases as the distance from the flashing source increases, the range of visibility can be typically a few hundred metres, depending on weather conditions. The attractiveness of a firefly is usually linked to the brightness of its flashes and the timing accuracy of its flashing patterns.

 Based on the above characteristics, Xin-She Yang developed in 2008 the firefly algorithm (FA) (Yang 2008, Yang 2010a). FA uses a nonlinear system by combing the exponential decay of light absorption and inverse-square law of light variation with distance. In the FA, the main algorithmic equation for the position $\x_i$ (as a solution vector to a problem) is
\be  \x_i^{t+1} =\x_i^t + \beta_0 e^{-\gamma r^2_{ij} }
(\x_j^t-\x_i^t) + \alpha \; \ff{\epsilon}_i^t, \label{FA-equ-100} \ee
where $\alpha$ is a scaling factor controlling the step sizes of the random walks,
while $\gamma$ is
a scale-dependent parameter controlling the visibility of the fireflies (and thus
search modes). In addition, $\beta_0$ is the attractiveness constant when the
distance between two fireflies is zero (i.e., $r_{ij}=0$). This system
is a nonlinear system, which may lead to rich characteristics in terms of
algorithmic behaviour.

Since the brightness of a firefly is associated with the objective landscape
with its position as the indicator, the attractiveness of a firefly seen by others, depending on their relative positions and relative brightness. Thus, the beauty is in the eye of the beholder. Consequently, a pair comparison is needed for comparing all fireflies. The main steps of FA can be summarized as the pseudocode in Algorithm~\ref{Alg-1}.

\begin{algorithm}
\hrule
Initialize all the parameters $\alpha, \beta, \gamma, n$\;
Initialize a population of $n$ firefies\;
Determine the light intensity/fitness at $\x_i$ by $f(\x_i)$\;
\While{$t<$ MaxGeneration}{
\For{All fireflies ($i=1:n$)}
{\For{All other fireflies ($j=1:n$) with $i \ne j$ (inner loop)}{
\If{Firefly $j$ is better/brighter than $i$}
{Move firefly $i$ towards $j$ using Eq.(\ref{FA-equ-100})\;}}
Evaluate the new solution\;
Accept the new solution if better\; }
Rank and update the best solution found\;
Update iteration counter $t \gets t+1$\;
Reduce $\alpha$ (randomness strength) by a factor\;
} \hrule
\caption{Firefly algorithm. \label{Alg-1}}
\end{algorithm}
It is worth pointing out that $\alpha$ is a parameter controlling the strength of the randomness or perturbations in FA. The randomness should be gradually reduced to speed up the overall convergence. Therefore, we can use
\be \alpha=\alpha_0 \delta^t, \ee
where $\alpha_0$ is the initial value and $0<\delta<1$ is a reduction factor. In most cases, we can use $\delta=0.9$ to $0.99$, depending on the type of problems and the desired quality of solutions.

If we look at Eq.(\ref{FA-equ-100}) closely, we can see that $\gamma$ is an important scaling parameter. At one extreme, we can set $\gamma=0$, which means that there is no exponential decay and thus the visibility is very high (all fireflies can see each other).
At the other extreme, when $\gamma \gg 1$, then the visibility range is very short.
Fireflies are essentially flying in a dense fog and they cannot see each other. Thus, each firefly flies independently and randomly. Therefore, a good value of $\gamma$ should be linked to the scale or limits of the design variables so that the fireflies within a range are visible to each other. This range is determined by
\be L=\frac{1}{\sqrt{\gamma}}, \ee
where $L$ the typical size of the search domain or the radius of a typical mode shape in the objective landscape. If there is no prior knowledge about its possible scale, we can start with $\gamma=1$ for most problems.

In fact, since FA is a nonlinear system, it has the ability to automatically
subdivide the whole swarm into multiple subswarms. This is because short-distance
attraction is stronger than long-distance attraction, and the division of
swarm is related to the mean range of attractiveness variations.
After division into multi-swarms, each subswarm can potentially swarm
around a local mode. Consequently, FA is naturally suitable for multimodal
optimization problems. Furthermore, there is no explicit use of the best solution $\ff{g}^*$,
thus selection is through the comparison of relative brightness
according to the rule of `beauty is in the eye of the beholder'.

It is worth pointing out that FA has some significant differences from PSO.
Firstly, FA is nonlinear, while PSO is linear. Secondly, FA has an ability of
multi-swarming, while PSO cannot. Thirdly, PSO uses velocities
(and thus have some drawbacks), while FA does not use velocities.
Finally, FA has some scaling control by using $\gamma$,
while PSO has no scaling control. All these differences enable FA to
search the design spaces more effectively for multimodal objective landscapes.

FA has been applied to a diverse range of applications and has been extended to multiobjective optimization and hybridization with other algorithms (Yang 2014a, Yang 2014b).

\subsection{Cuckoo Search}
\index{cuckoo search}

In the natural world, among 141 cuckoo species, 59 species engage the so-called obligate brood parasitism (Davies 2011). These cuckoo species do not build their own nests and they lay eggs in the nests of host birds such as warblers. Sometimes, host birds can spot the alien eggs laid by cuckoos and thus can get rid of the eggs or abandon the nest by flying away to build a new nest in a new location so as to reduce the possibility of raising an alien cuckoo chick. The eggs of cuckoos can be sufficiently similar to eggs of host birds in terms the size, color and texture so as to increase the survival probability of cuckoo eggs. In reality, about 1/5 to 1/4 of eggs laid by cuckoos will be discovered and abandoned by hosts.
In fact, there is an arms race between cuckoo species and host species, forming an interesting cuckoo-host species co-evolution system. \index{co-evolution}

Based the above characteristics, Yang and Deb developed in 2009 the cuckoo search (CS) algorithm (Yang and Deb 2009).
CS uses a combination of both local and global search capabilities, controlled by a discovery  probability $p_a$.
There are two algorithmic equations in CS, and one equation is
\be \x_i^{t+1}=\x_i^t +\alpha s \otimes H(p_a-\epsilon) \otimes (\x_j^t-\x_k^t), \label{CS-equ1} \ee
where $\x_j^t$ and $\x_k^t$ are two different solutions selected randomly by random permutation,
$H(u)$ is a Heaviside function, $\epsilon$ is a random number drawn from a uniform distribution, and $s$ is the step size. This step is primarily local, though it can become global search if $s$ is large enough. However, the main global search mechanism is realized by the other equation with L\'evy flights:
\be \x_i^{t+1}=\x_i^t+\alpha L(s,\lam), \label{CS-equ2} \ee
where the L\'evy flights are simulated (or drawn random numbers) by drawing random numbers from a L\'evy distribution
\be L(s,\lam) \sim \frac{\lam \Gamma(\lam) \sin (\pi \lam/2)}{\pi}
\frac{1}{s^{1+\lam}}, \quad (s \gg 0). \ee
Here $\alpha>0$ is the step size scaling factor.

By looking at the equations in CS carefully, we can clearly see
that CS is a nonlinear system due to the Heaviside function,
discovery probability and L\'evy flights. There is no explicit use of global best $\ff{g}^*$,
but selection of the best solutions is by ranking and elitism where the current best
is passed onto the next generation. In addition, the use of  L\'evy flights
can enhance the search capability because a fraction of steps generated by L\'evy flights
are larger than those used in Gaussian. Thus, the search steps in CS
are heavy-tailed (Pavlyukevich 2007, Reynolds and Rhodes 2009).
Consequently, CS can be very effective for nonlinear optimization problems and multiobjective optimization (Gandomi et al. 2013, Yang and Deb 2013; Yang 2014a, Yildiz 2013). A relatively comprehensive literature review of cuckoo search has been carried out by Yang and Deb (2014).

\subsection{Other Algorithms}

As we mentioned earlier, the literature is expanding and more social algorithms are being developed by researchers, but we will not introduce more algorithms here. Instead, we will focus on summarizing the key characteristics of social algorithms and other population-based algorithms so as to gain a deeper understanding of these algorithms.

\section{Algorithm Analysis and Insight}

\subsection{Characteristics of Social Algorithms }

Though different social algorithms have different characteristics and inspiration from nature, they do share some common features. Now let us look at these algorithms in terms of  their basic steps, search characteristics
and algorithm dynamics.
\begin{itemize}
\item All social algorithms use a population of multiple agents (e.g., particles, ants,
bats, cuckoos, fireflies, bees, etc.), each agent corresponds to a solution vector. Among the population, there is often the best solution $\ff{g}^*$ in terms of objective fitness.
Different solutions in a population represent both diversity and different fitness.

\item  The evolution of the population is often achieved by some operators (e.g., mutation by a vector or by randomization), often in terms of some algorithmic formulas or equations. Such evolution is typically iterative, leading to evolution of solutions with different properties. When all solutions become sufficiently similar, the system can be considered as converged.

\item All algorithms try to carry out some sort of both local and global search. If the search is mainly local, it increases the probability of getting stuck locally. If the search focuses too much on global moves, it will slow down the convergence. Different algorithms may use different amount of randomization and different portion of moves for local or global search so as to balance exploitation and exploration (Blum and Roli 2001).

\item Selection of the better or best solutions is carried out by the `survival of the fittest' or simply elitism so that the best solution $\ff{g}^*$ is kept in the population in the next generation. Such selection essentially acts a driving force to drive the diverse population into a converged population with reduced diversity but with a more organized structure.

\end{itemize}
These basic components, characteritics and their properties can be summarized in Table \ref{table-char}.
\begin{table}
\begin{center}
\caption{Characteristics and properties of social algorithms. \label{table-char} }
\begin{tabular}{|l|l|l|}
\hline
Components/characteristics & Role or Properties \\
\hline \hline
Population (multi-agents) & Diversity and sampling \\
\hline
 Randomization/perturbations &  Exploration and escape local optima \\
\hline
 Selection and elitism & Exploitation and driving force for convergence \\
\hline
 Algorithmic equations & Iterative evolution of solutions \\
\hline
\end{tabular}
\end{center}
\end{table}

\subsection{No Free Lunch Theorems}
\index{no-free-lunch theorem}

Though there are many algorithms in the literature, different algorithms can have different
advantages and disadvantages and thus some algorithms are more suitable to solve certain types of problems than others. However, it is worth pointing out that
there is no single algorithm that can be most efficient to solve all types of problems as dictated by the no-free-lunch (NFL) theorems (Wolpert and Macready, 1997).
Even the no-free-lunch theorems hold under certain conditions, but these conditions may not be rigorously true for actual algorithms. For example, one condition for proving these theorems is the so-called no-revisiting condition. That is, the points during iterations form a path, and these points are distinct and will not be visited exactly again, though their nearby neighbourhood can be revisited. This condition is not strictly valid because almost all algorithms for continuous optimization will revisit some of their points in history. Such minor violation of assumptions can potentially leave room for free lunches.
It has also been shown that under the right conditions such as co-evolution, certain algorithms can be more effective (Wolpert and Macready 2005).

In addition, a comprehensive review by T. Joyce and J.M. Herrmann
on no-free-lunch (NFL) theorems (Joyce and Herrmann 2018), free lunches may exist for a finite set of problems, especially those algorithms that can exploit the objective landscape structure and knowledge of optimization problems to be solved. If the performance is not averaged over all {\it possible} problems, then free lunches can exist. In fact, for a given finite set of problems and a finite set of algorithms, the comparison is essentially equivalent to a zero-sum ranking problem. In this case, some algorithms can perform better than others for solving a {\it certain type} of problems. In fact, almost all research papers published about comparison of algorithms use a few algorithms and a finite set (usually under 100 benchmarks), such comparisons are essentially ranking. However, it is worth pointing out that for a finite set of benchmarks, the conclusions (e.g., ranking) obtained can only apply for that set of benchmarks, they may not be valid for other sets of benchmarks and the conclusions can be significantly different. If interpreted in this sense, such comparison studies and their conclusions are consistent with NFL theorems.

\section{Future Directions}

The research area of social algorithms is very active, and there are many hot topics for further research directions concerning these algorithms. Here we highlight a few:
\begin{itemize}

\item {\bf Theoretical framework}: Though there are many studies concerning the implementations and applications of social algorithms, mathematical analysis of such algorithms lag behind. There is a strong need to build a theoretical framework to analyze these algorithms mathematically so as to gain in-depth understanding (He et al. 2017). For example, it is not clear how local rules can lead to the rise of self-organized structure in algorithms. More generally, it still lacks key understanding about the
    rise of social intelligence and swarm intelligence in a multi-agent system and their
    exact conditions.

      \index{parameter tuning}
\item {\bf Parameter tuning}: Almost all algorithms have algorithm-dependent parameters, and the performance of an algorithm is largely influenced by its parameter setting. Some parameters may have stronger influence than others (Eiben and Smit 2011). Therefore, proper parameter tuning and sensitivity analysis are needed to tune algorithms to their best. However, such tuning is largely done by trial and error in combination with the empirical observations. How to tune them quickly and automatically is still an open question.

\item {\bf Large-scale applications}: Despite the success of social algorithms and their diverse applications, most studies in the literature have concerned problems of moderate size with the number of variables up to a few hundred at most. In real-world applications, there may be thousands and even millions of design variables, it is not clear yet how these algorithms can be scaled up to solve such large-scale problems.
    In addition, though social algorithms have been applied to solve combinatorial problems such as scheduling and the travelling salesman problem with promising results, these problems are typically non-deterministic polynomial-time (NP) hard, and thus for larger problem sizes, they can be very challenging to solve. Researchers are not sure how to modify exist social algorithms to cope with such challenges.

     \index{hybrid algorithm} \index{co-evolutionary algorithm}
\item {\bf Hybrid and co-evolutionary algorithms}: The algorithms we have covered here are algorithms that are `pure' and `standard' in the sense that they have not been heavily modified by hybridizing with others. Both empirical observations and studies show that the combination of the advantages from two or more different algorithms can produce a better hybrid, which can use the distinct advantages of its component algorithms and potentially avoid their drawbacks. In addition, it is possible to build a proper algorithm system to allow a few algorithms to co-evolve to obtain an overall better performance.

    Though NFL theorems may hold for simple algorithms, it has been shown that there can be free lunches for co-evolutionary algorithms (Wolpert and Macready 2005). Therefore, future research can focus on figuring out how to assemble different algorithms into an efficient co-evolutionary system and then tune the system to its best.

\item {\bf Self-Adaptive and self-evolving  algorithms}: Sometimes, the parameters in an algorithm can vary to suit for different types of problems. In addition to parameter tuning, such parameter adaptivity
    can be advantageous. There are some basic form of adaptive algorithms in the literature and they mainly use random variations of parameters in a fixed range. Ideally, an algorithm should be self-adaptive and be able to automatically tune itself to suit for a given type of problems without much supervision from the users (Yang et al. 2013). Such algorithms should also be able to evolve by learning from their past performance histories. The ultimate aim for researchers is to build a set of self-adaptive, self-tuning, self-learning and self-evolving algorithms that can solve a diverse range of real-world applications efficiently and quickly in practice.  \index{self-adaptivity} \index{self-evolving algorithm}

\end{itemize}

Social algorithms have become a powerful tool set for solving optimization problems and their studies form an active area of research. It is hoped that this work can inspire more research concerning the above important topics.


\end{document}